# EVALUATION OF PERFORMANCE MEASURES FOR CLASSIFIERS COMPARISON

**Vincent Labatut**
Galatasaray University, Computer Science Department, Istanbul, Turkey
vlabatut@gsu.edu.tr

**Hocine Cherifi**
University of Burgundy, LE2I UMR CNRS 5158, Dijon, France
hocine.cherifi@u-bourgogne.fr

**ABSTRACT**
The selection of the best classification algorithm for a given dataset is a very widespread problem, occuring each time one has to choose a classifier to solve a real-world problem. It is also a complex task with many important methodological decisions to make. Among those, one of the most crucial is the choice of an appropriate measure in order to properly assess the classification performance and rank the algorithms. In this article, we focus on this specific task. We present the most popular measures and compare their behavior through discrimination plots. We then discuss their properties from a more theoretical perspective. It turns out several of them are equivalent for classifiers comparison purposes. Futhermore. they can also lead to interpretation problems. Among the numerous measures proposed over the years, it appears that the classical overall success rate and marginal rates are the more suitable for classifier comparison task.

**Keywords:** Classification, Accuracy Measure, Classifier Comparison, Discrimination Plot.

## 1 INTRODUCTION

The comparison of classification algorithms is a complex and open problem. First, the notion of performance can be defined in many ways: accuracy, speed, cost, readability, etc. Second, an appropriate tool is necessary to quantify this performance. Third, a consistent method must be selected to compare the measured values.

As performance is most of the time expressed in terms of accuracy, we focus on this point in this work. The number of accuracy measures appearing in the classification literature is extremely large. Some were specifically designed to compare classifiers , but most were initially defined for other purposes, such as measuring the association between two random variables [2], the agreement between two raters [3] or the similarity between two sets [4]. Furthermore, the same measure may have been independently developed by different authors, at different times, in different domains, for different purposes, leading to very confusing typology and terminology. Besides its purpose or name, what characterizes a measure is the definition of the concept of accuracy it relies on. Most measures are designed to focus on a specific aspect of the overall classification results [5]. This leads to measures with different interpretations, and some do not even have any clear interpretation. Finally, the measures may also differ in the nature of the situations they can handle [6]. They can be designed for binary (only two classes) or multiclass (more than two classes) problems. They can be dedicated to mutually exclusive (one instance belongs to exactly one class) or overlapping classes (one instance can belong to several classes) situations. Some expect the classifier to output a discrete score (Boolean classifiers), whereas other can take advantage of the additional information conveyed by a real-valued score (probabilistic or fuzzy classifiers). One can also oppose flat (all classes on the same level) and hierarchical classification (a set of classes at a lower level constitutes a class at a higher level). Finally, some measures are sensitive to the sampling design used to retrieve the test data [7].

Many different measures exist, but yet, there is no such thing as a perfect measure, which would be the best in every situation [8]: an appropriate measure must be chosen according to the classification context and objectives. Because of the overwhelming number of measures and of their heterogeneity, choosing the most adapted one is a difficult problem. Moreover, it is not always clear what the measures properties are, either because they were never rigorously studied, or because specialists do not agree on the question (e.g. the question of chance-correction [9]). Maybe for these reasons, authors very often select an accuracy measure by relying on the tradition or consensus observed in their field. The point is then more to use the same measure than their peers rather than the most appropriate one.

In this work, we reduce the complexity of choosing an accuracy measure by restraining our analysis to a very specific but widespread, situation. We discuss the case where one wants to select the best classification algorithm to process a given data set [10]. An appropriate way to perform this task would be to study the data properties first, then to select a suitable classification algorithm and determine the most appropriate parameter values, and finally to use it to build the classifier. But not everyone has the statistical expertise required to perform this analytic work. Therefore, in practice, the most popular method consists in sampling a training set from the considered population, building various classifiers with different classification algorithms and/or parameters, and then comparing their performances empirically on some test sets sampled from the same population. Finally, the classifier with highest performance is selected and used on the rest of the population.

We will not address the question of the method used to compare performances. Instead, we will discuss the existing accuracy measures and their relevance to our specific context. We will be focusing on comparing basic classifiers, outputting discrete scores for flat mutually-exclusive classes. Throughout this paper, we will make the following assumptions linked to our context. First, as we want to discriminate some classifiers, if two measures rank them similarly, we consider these measures as equivalent, even if they do not return the exact same accuracy values. Second, since we compare some classifiers on the same dataset, the class proportions in the processed data are fixed.

In the next section, we review the works dealing with similar problems. We then introduce the notations used in the rest of the paper in section 3. In section 4, we review the main measures used as accuracy measures in the classification literature. In section 5, we compare them empirically, by considering some typical cases. In section 6, we introduce the notion of discrimination plot to compare and analyze the behavior of the measures. Finally, in section 7, we compare their functional properties and discuss their relevance relatively to our specific case.

## 2 RELATED WORKS

Several previous works already compared various measures, but with different purposes or methods. In [11], Congalton described the various aspects of accuracy assessment and compared a few measures in terms of functional traits. However, the focus is rather on estimating the quality of a single classifier than on comparing several of them. In other words, the discussion concerns whether or not the value of a given measure is close to the studied classifier actual accuracy, and not on the ability of this measure to discriminate between classifiers.

Ling *et al.* defined the notions of consistency and discriminancy to compare measures [12]. They stated two measures are consistent if they always discriminate algorithms similarly. A measure is more discriminant than the other if it is the only one (of the two) sensitive to differences in the processed data. The authors use these concepts to compare 2 widespread measures. The notion of consistency fits the previous definition of measure equivalence we adopted in our specific context. However, Ling *et al.*'s focus is on real-valued output scores and binary classification problems.

In [13], Flach compared 7 measures through the use of ROC plots. He studied how these measures behave when varying the classes relative proportions in the dataset. For this purpose, he considered the isometrics of a given measure (i.e. the zones of the ROC space for which the measure returns the same value), and investigated how changes in the class proportions affect them. He defined the equivalence of two measures in the context of classifiers comparison in a way relatively similar to Ling *et al.*'s consistency [12]. His work also focused on binary problems.

Sokolova & Lapalme considered 24 measures, on both binary and multiclass problems (and others) [6]. They studied the sensitivity of these measures to specific changes in the classified dataset properties. Using the same general idea than Flach [13] (isometrics), they developed the notion of invariance, by identifying the changes in the confusion matrix which did not affect the measure value. Note they focused on class-specific changes. The measures were compared in terms of invariance: two measures are said to be similar if they are invariant to the same modifications. This is stricter than what we need in our context, since some modification might change the accuracy but not the algorithms relative ranking.

In [14], Albatineh *et. al* performed an analytical study of 28 accuracy measures. They considered these measures in the context of cluster analysis accuracy assessment, but the $2 \times 2$ confusion matrices they analyzed are similar to those obtained for binary classification problems. They showed many of the considered measures are equivalent (i.e. return the same values) when a correction for chance (cf. section 4.6) is applied. Besides the fact the authors focus on binary problems, this work also differs from ours because of the much stricter notion of equivalence: two measures can provide different values but rank classifiers similarly. Moreover, the relevance of chance correction has yet to be discussed in our context.

By opposition to the previous analytical works, a number of authors adopted an empirical approach. The general idea is to apply several classifiers to a selection of real-world data sets, and to process their accuracy through various measures. These are then compared in terms of correlation. Caruana & Niculescu-Mizil adopted this method to compare 9 accuracy measures [15], but their focus was on binary classification problems, and classifiers able to output real-valued scores (by opposition to the discrete scores we treat here). Liu *et al.* [16] and and Ferri *et al.* [17] considered 34 and 18 measures, respectively, for both binary and multiclass problems (amongst others). The main limitation with these studies is they either use data coming from a single applicative domain (such as remote sensing in [16]), or rely on a small number of datasets (7 in [15] and 30 in [17]). In both cases, this prevents a proper generalization of the obtained observations. Ferri *et al.* completed their empirical analysis by studying the effect of various types of noise on the measures, through randomly generated data. However their goal was more to characterize the measures sensitivity than to compare them directly.

## 3  NOTATIONS AND TERMINOLOGY

Consider the problem of estimating $k$ classes for a test set containing $n$ instances. The true classes are noted $C_i$, whereas the *estimated* classes, as defined by the considered classifier, are noted $\hat{C}_i$ ( $1 \leq i \leq k$ ). The proportion of instances belonging to class $C_i$ in the dataset is noted $\pi_i$.

Most measures are not processed directly from the raw classifier outputs, but from the *confusion matrix* built from these results. This matrix represents how the instances are distributed over estimated (rows) and true (columns) classes.

**Table 1:** a general confusion matrix.

|  | $C_1$ | ... | $C_k$ |
|---|---|---|---|
| $\hat{C}_1$ | $p_{11}$ | ... | $p_{1k}$ |
| $\vdots$ | $\vdots$ | $\ddots$ | $\vdots$ |
| $\hat{C}_k$ | $p_{k1}$ | ... | $p_{kk}$ |

In Table 1, the terms $p_{ij}$ ( $1 \leq i, j \leq k$ ) correspond to the proportion of instances estimated to be in class number $i$ by the classifier (i.e. $\hat{C}_i$), when they actually belong to class number $j$ (i.e. $C_j$). Consequently, diagonal terms ( $i = j$ ) correspond to correctly classified instances, whereas off-diagonal terms ( $i \neq j$ ) represent incorrectly classified ones. Note some authors invert estimated and true classes, resulting in a transposed matrix [13, 18].

The sums of the confusion matrix elements over row $i$ and column $j$ are noted $p_{i+}$ and $p_{+j}$, respectively, so we have $p_{+j} = \pi_j$.

When considering one class $i$ in particular, one may distinguish four types of instances: true positives (TP) and false positives (FP) are instances correctly and incorrectly classified as $\hat{C}_i$, whereas true negatives (TN) and false negatives (FN) are instances correctly and incorrectly not classified as $\hat{C}_i$, respectively. The corresponding proportions are defined as $p_{TP} = p_{ii}$, $p_{FN} = p_{+i} - p_{ii}$, $p_{FP} = p_{i+} - p_{ii}$ and $p_{TN} = 1 - p_{TP} - p_{FP} - p_{FN}$, respectively.

Note some authors prefer to define the confusion matrix in terms of counts rather than proportions, using values of the form $n_{ij} = np_{ij}$. Since using proportions is generally more convenient when expressing the accuracy measures, we prefer to use this notation in this article.

## 4  SELECTED MEASURES

In this section, we describe formally the most widespread measures used to compare classifiers in the literature. These include association measures, various measures based on marginal rates of the confusion matrix, and chance-corrected agreement coefficients. Since these can be described according to many traits, there are as many typologies as authors. In this article, we will mainly oppose class-specific measures, i.e. those designed to assess the accuracy of a single class, and multiclass measures, able to assess the overall classifier accuracy. Note the class-specific ones generally correspond to measures defined for binary problems and applied on multiclass ones.

### 4.1  Nominal Association Measures

A measure of association is a numerical index, a single number, which describes the strength or magnitude of a relationship. Many association measures were used to assess classification accuracy, such as: chi-square-based measures ($\Phi$ coefficient, Pearson's $C$, Cramer's $V$, etc. [2]), Yule's coefficients, Matthew's correlation coefficient, Proportional reduction in error measures (Goodman & Kruskal's $\lambda$ and $\pi$, Theil's uncertainty coefficient, etc.), mutual information-based measures [19] and others. Association measures quantify how predictable a variable is when knowing the other one. They have been applied to classification accuracy assessment by considering these variables are defined by the distributions of instances over the true and estimated classes, respectively.

In our context, we consider the distribution of instances over estimated classes, and want to measure how much similar it is to their distribution over the true classes. The relationship assessed by an association measure is more general [2], since a high level of association only means it is possible to predict estimated classes when knowing the true ones (and vice-versa). In other terms, a high association does not necessary correspond to a match between estimated and true classes. For instance, if one considers a binary classification problem, both perfect classification and perfect misclassification give the same maximal association value.

**Table 2:** confusion matrix displaying a case of perfect misclassification leading to a maximal association measure.

|   | $C_1$ | $C_2$ | $C_3$ |
|---|---|---|---|
| $\hat{C}_1$ | 0.00 | 0.00 | 0.33 |
| $\hat{C}_2$ | 0.33 | 0.00 | 0.00 |
| $\hat{C}_3$ | 0.00 | 0.34 | 0.00 |

Consequently, a confusion matrix can convey both a low accuracy and a high association at the same time (as shown in Table 2), which makes association measures unsuitable for accuracy assessment.

### 4.2 Overall Success Rate

Certainly the most popular measure for classification accuracy [20], the *overall success rate* is defined as the trace of the confusion matrix:

$$OSR = \sum_{i=1}^{k} p_{ii} \qquad (1)$$

This measure is multiclass, symmetrical, and ranges from 0 (perfect misclassification) to 1 (perfect classification). Its popularity is certainly due to its simplicity, not only in terms of processing but also of interpretation, since it corresponds to the observed proportion of correctly classified instances.

### 4.3 Marginal Rates

We gather under the term *marginal rates* a number of widely spread asymmetric class-specific measures. The *TP Rate* and *TN Rate* are both reference-oriented, i.e. they consider the confusion matrix columns (true classes). The former is also called sensitivity [20], producer's accuracy [11] and Dice's asymmetric index [21]. The latter is alternatively called specificity [20].

$$TPR_i = p_{TP} / (p_{TP} + p_{FN}) \qquad (2)$$
$$TNR_i = p_{TN} / (p_{TN} + p_{FP}) \qquad (3)$$

The estimation-oriented measures, which focus on the confusion matrix rows (estimated classes), are the *Positive Predictive Value* (PPV) and *Negative Predictive Value* (NPV) [20]. The former is also called precision [20], user's accuracy [11] and Dice's association index [21].

$$PPV_i = p_{TP} / (p_{TP} + p_{FP}) \qquad (4)$$
$$NPV_i = p_{TN} / (p_{TN} + p_{FN}) \qquad (5)$$

TNR and PPV are related to *type I error* (FP) whereas TPR and NPV are related to *type II error* (FN). All four measures range from 0 to 1, and their interpretation is straightforward. TPR (resp. TNR) corresponds to the proportion of instances belonging (resp. not belonging) to the considered class and actually classified as such. PPV (resp. NPV) corresponds to the proportion of instances predicted to belong (resp. not to belong) to the considered class, and which indeed do (resp. do not).

Finally, note some authors use the complements of these measures. For instance, the *False Positive Rate* $FPR_i = 1 - TNR_i$ is also called fallout [22] or false alarm rate [23], and is notably used to build ROC curves [20].

### 4.4 F-measure and Jaccard Coefficient

The *F-measure* corresponds to the harmonic mean of PPV and TPR [20], therefore it is class-specific and symmetric. It is also known as F-score [15], Sørensen's similarity coefficient [24], Dice's coincidence index [21] and Hellden's mean accuracy index [25]:

$$F_i = 2 \frac{PPV_i \times TPR_i}{PPV_i + TPR_i} = \frac{2 p_{TP}}{2 p_{TP} + p_{FN} + p_{FP}} \qquad (6)$$

It can be interpreted as a measure of overlapping between the true and estimated classes (other instances, i.e. TN, are ignored), ranging from 0 (no overlap at all to 1 (complete overlap).

The measure known as *Jaccard's coefficient of community* was initially defined to compare sets [4], too. It is a class-specific symmetric measure defined as:

$$JCC_i = p_{TP} / (p_{TP} + p_{FP} + p_{FN}) \qquad (7)$$

It is alternatively called Short's measure [26]. For a given class, it can be interpreted as the ratio of the estimated and true classes intersection to their union (in terms of set cardinality). It ranges from 0 (no overlap) to 1 (complete overlap). It is related to the F-measure [27]: $JCC_i = F_i / (2 - F_i)$, which is why we describe it in the same section.

### 4.5 Classification Success Index

The *Individual Classification Success Index* (ICSI), is a class-specific symmetric measure defined for classification assessment purpose [1]:

$$\begin{aligned} ICSI_i &= 1 - (1 - PPV_i + 1 - TPR_i) \\ &= PPV_i + TPR_i - 1 \end{aligned} \qquad (8)$$

The terms $1 - PPV_i$ and $1 - TPR_i$ correspond to the proportions of type I and II errors for the considered class, respectively. ICSI is hence one minus the sum of these errors. It ranges from –1 (both errors are maximal, i.e. 1) to 1 (both errors are minimal, i.e. 0), but the value 0 does not have any clear meaning. The measure is symmetric, and linearly related to the arithmetic mean of TPR and PPV, which is itself called *Kulczynski's measure* [28].

The *Classification Success Index* (CSI) is an overall measure defined simply by averaging ICSI over all classes [1].

### 4.6 Agreement Coefficients

A family of chance-corrected inter-rater agreement coefficients has been widely used in the context of classification accuracy assessment. It relies on the following general formula:

$$A = (P_o - P_e) / (1 - P_e) \qquad (9)$$

Where $P_o$ and $P_e$ are the observed and expected agreements, respectively. The idea is to consider the observed agreement as the result of an intended agreement and a chance agreement. In order to get the intended agreement, one must estimate the chance agreement and remove it from the observed one.

Most authors use $P_o = OSR$, but disagree on how the chance agreement should be formally defined, leading to different estimations of $P_e$. For his popular *kappa coefficient* (CKC), Cohen used the product of the confusion matrix marginal proportions [3]: $P_e = \sum_i p_{i+} p_{+i}$. Scott's *pi coefficient* (SPC) relies instead on the class proportions measured on the whole data set (or its estimation), noted $p_i$ [29]: $P_e = \sum_i (p_i)^2$. Various authors, including Maxwell for his *Random Error* (MRE) [30], made the assumption classes are evenly distributed: $P_e = 1/k$.

The problems of assessing inter-rater agreement and classifier accuracy are slightly different though. Indeed, in the former, the true class distribution is unknown, whereas in the latter it is completely known. Both raters are considered as equivalent and interchangeable, in the sense they are both trying to estimate the true classes. On the contrary, in our case, the classes estimated by the classifier are evaluated relatively to the true classes. The correction for chance strategies presented above are defined in function of this specific trait of the inter-rater agreement problem. They might not be relevant in our situation.

### 4.7 Ground Truth Index

Türk's *Ground Truth Index* (GTI) is another chance-corrected measure, but this one was defined specially for classification accuracy assessment [18]. Türk supposes the classifier has two components: one is always correct, and the other classify randomly. For a given class $C_j$, a proportion $\theta_j$ of the instances are supposed to be classified by the infallible classifier, and therefore put in $\hat{C}_j$. The remaining instances (i.e. a proportion $b_j = 1 - \theta_j$) are distributed by the random classifier over all estimated classes (including $\hat{C}_j$) with a probability $a_i$ for $\hat{C}_i$. In other words, according to this model, each off-diagonal term of the confusion matrix can be written as a product of the form $p_{ij} = a_i b_j$ ($i \neq j$). This corresponds to the hypothesis of quasi-independence of non-diagonals of Goodman, whose iterative proportional fitting method allows estimating $a_i$ and $b_j$.

Türk based his GTI $\theta$ on the general formula of Eq. (9), but unlike the previously presented agreement coefficients, he uses $P_o = TPR_i$ and $P_e = a_i$. He therefore designs a class-specific measure, corresponding to a chance-corrected version of the TPR. It is interpreted as the proportion of instances the classifier will always classify correctly, even when processing other data. The way this measure handles chance correction is more adapted to classification than the agreement coefficients [27]. However, it has several limitations regarding the processed data: it cannot be used with less than three classes, or on perfectly classified data, and most of all it relies on the quasi-independence hypothesis. This condition is extremely rarely met in practice (e.g. less than 10% of the real-world cases considered in [16]). For this reason we will not retain the GT index in our study.

Finally, Table 3 displays the measures selected to be studied more thoroughly in the rest of this article, with their main properties

**Table 3:** selected accuracy measures and their main properties: focus (either multiclass –MC– or class-specific –CS), chance-corrected, symmetrical and range.

| Name | Focus | Ch. | Sym. | Range |
|---|---|---|---|---|
| OSR | MC | No | Yes | $[0;1]$ |
| TPR | CS | No | No | $[0;1]$ |
| TNR | CS | No | No | $[0;1]$ |
| PPV | CS | No | No | $[0;1]$ |
| NPV | CS | No | No | $[0;1]$ |
| F-meas. | CS | No | No | $[0;1]$ |
| JCC | CS | No | No | $[0;1]$ |
| ICSI | CS | No | No | $[-1;1]$ |
| CSI | MC | No | No | $[-1;1]$ |
| CKP | MC | Yes | Yes | $[-\infty;1]$ |
| SPC | MC | Yes | Yes | $[-\infty;1]$ |
| MRE | MC | Yes | Yes | $\left[\frac{-1}{k-1};1\right]$ |

## 5 CASE STUDIES

In this section, we discuss the results obtained on a few confusion matrices in order to analyze the properties and behavior of the measures reviewed in the previous section. We first consider extreme cases, i.e. perfect classification and misclassification. Then study a more realistic confusion matrix including a few classification errors.

### 5.1 Extreme Cases

All measures agree to consider diagonal confusion matrices such as the one presented in Table 4 as the result of a perfect classification. In this case, the classifier assigns each instance to its true class and the measure reaches its maximal value.

A perfect misclassification corresponds to a matrix whose trace is zero, as shown in Tables 2 and 5. In this case, the measures diverge. Their behavior depends on the way they consider the distribution of errors over the off-diagonal cells. OSR is not sensitive to this distribution since only the trace of the confusion matrix is considered. TPR, PPV, JCC and F-measure are not concerned neither, since having no TP automatically causes these measure to have a zero value. CSI consequently always reach its minimal value too, since it depends directly on TPR and PPV, and so does ICSI.

**Table 4:** A confusion matrix illustrating a case of perfect classification.

|   | $C_1$ | $C_2$ | $C_3$ |
|---|---|---|---|
| $\hat{C}_1$ | 0.33 | 0.00 | 0.00 |
| $\hat{C}_2$ | 0.00 | 0.34 | 0.00 |
| $\hat{C}_3$ | 0.00 | 0.00 | 0.33 |

The chance-corrected measures are affected according to the model of random agreement they are based upon. For the misclassification case depicted by Table 2, all of them have the same value 0.5. But for the misclassification case observed in Table 5, we obtain the following values: $CKC = -0.43$, $SPC = -0.61$ and $MRE = -0.50$. This is to compare with the previously cited measures, which do not discriminate these two cases of perfect misclassification.

**Table 5:** A confusion matrix illustrating a case of perfect misclassification.

|   | $C_1$ | $C_2$ | $C_3$ |
|---|---|---|---|
| $\hat{C}_1$ | 0.00 | 0.10 | 0.10 |
| $\hat{C}_2$ | 0.30 | 0.00 | 0.10 |
| $\hat{C}_3$ | 0.20 | 0.20 | 0.00 |

TNR and NPV react differently to the perfect misclassification case. Indeed, they are both related to the number of TN, which might not be zero even in case of perfect misclassification. In other words, provided each class is represented in the considered dataset, these measures cannot reach their minimal value for all classes in case of perfect misclassification. For instance, in Table 5 $TNR_1 = 0.6$ because 60% of the non-$C_1$ instances are not classified as $\hat{C}_1$.

### 5.2 Intermediate Cases

Let us consider the confusion matrix displayed in Table 6, whose associated accuracy values are given in Table 7. We focus on the marginal rates first, beginning with the first class. The very high

TPR indicates the classifier is good at classifying instances belonging to $C_1$ (91% of them are placed in $\hat{C}_1$), but not as much for instances belonging to other classes (lower TNR: only 79% of the non-$C_1$ instances are not put in $\hat{C}_1$). The predictive rates address the quality of the estimated classes, showing the classifier estimation is reliable for classes other than $\hat{C}_1$ (high NPV, 95% of the non-$\hat{C}_1$ instances actually do not belong to $C_1$), but not as much for $\hat{C}_1$ (lower PPV, only 68% of the instances in $\hat{C}_1$ actually belong to $C_1$).

**Table 6:** A confusion matrix illustrating an intermediate case of classification. The classifier is weaker for the second class.

|  | $C_1$ | $C_2$ | $C_3$ |
|---|---|---|---|
| $\hat{C}_1$ | 0.30 | 0.12 | 0.02 |
| $\hat{C}_2$ | 0.02 | 0.19 | 0.01 |
| $\hat{C}_3$ | 0.01 | 0.03 | 0.30 |

The third class is interesting, because the other values in its column are the same as for the first class, whereas those on its row are better (i.e. smaller). The first observation explains why its TPR and NPV are similar to those of the first class, and the second why its TNR and PPV are higher. In other words, the classifier is better as classifying instances not belonging to $C_3$ (higher TNR, 94% of the non-$C_3$ instances are not put in $\hat{C}_3$) and the estimated class $\hat{C}_3$ is much more reliable (higher PPV, 88% of the instances in $\hat{C}_3$ actually belong to $C_3$).

**Table 7:** accuracy values associated with the confusion matrix of Table 6.

|  | Cls.1 | Cls.2 | Cls.3 | Multi. |
|---|---|---|---|---|
| OSR | - | - | - | 0.79 |
| TPR | 0.91 | 0.56 | 0.91 | - |
| TNR | 0.79 | 0.95 | 0.94 | - |
| PPV | 0.68 | 0.86 | 0.88 | - |
| NPV | 0.95 | 0.81 | 0.95 | - |
| F-meas. | 0.78 | 0.68 | 0.90 | - |
| JCC | 0.64 | 0.51 | 0.81 | - |
| (I)CSI | 0.59 | 0.42 | 0.79 | 0.60 |
| CKP | - | - | - | 0.69 |
| SPC | - | - | - | 0.68 |
| MRE | - | - | - | 0.69 |

Finally, let us consider the second class, which is clearly the weakest for this classifier. The low TPR indicates the classifier has trouble recognizing all instances from $C_2$ (only 56% are correctly classified). However, the other measures are relatively high: it manages not putting in $\hat{C}_2$ 95% of the non-$C_2$ instances (TNR), and its estimations for $\hat{C}_2$ and the other classes are reliable: 86% of the instances put in $\hat{C}_2$ actually belong to $C_2$ (PPV) and 81% of the instances not put in $\hat{C}_2$ actually do not belong to $C_2$ (NPV).

The other class-specific measure (F-measure, JCC and CSI) corroborates the conclusions drawn for the marginal rates. They indicate the classifier is better on the third, then first, and second classes. Intuitively, one can deduce from these comments the classifier confuses some of the second class instances and incorrectly put them in the first one.

Now suppose we want to consider the performance of another classifier on the same data: then only the repartitions of instances in each column can change (the repartitions cannot change along the rows, since these depend on the dataset). Let us assume the new classifier is perfect on $C_1$ and $C_3$, but distributes $C_2$ instances uniformly in $\hat{C}_1$, $\hat{C}_2$ and $\hat{C}_3$, as shown in Table 8. We obtain the following values for the multiclass measures: $OSR = 0.78$, $ICSI = 0.62$, $CKP$, $SPC$, $MRE = 0.67$. Interestingly, all multiclass measures consider the first classifier as better, except ICSI. This is due to the fact the average decrease in TPR observed for the second classifier relatively to the first is compensated by the PPV increase. This illustrates the fact all measures do not necessarily rank the classifiers similarly. Note that from the results reported in Table 7 one could have thought the contrary.

**Table 8:** confusion matrix displaying the classification results of a second classifier.

|  | $C_1$ | $C_2$ | $C_3$ |
|---|---|---|---|
| $\hat{C}_1$ | 0,33 | 0,11 | 0,00 |
| $\hat{C}_2$ | 0,00 | 0,12 | 0,00 |
| $\hat{C}_3$ | 0,00 | 0,11 | 0,33 |

## 6 SENSITIVITY TO MATRIX CHANGES

Since measures possibly discriminate classifiers differently, we now focus on the nature of this disagreement. We study three points likely to affect the accuracy measures: the classifier distribution of error, the dataset class proportions and the number of classes.

## 6.1 Methods

We first give an overview of our methods, and then focus on its different steps. We generate two series of matrices with various error distribution and fixed class proportions. We compute the accuracy according to every measure under study. For a given measure we then consider all possible pairs of matrices obtained by associating one matrix from the first series to one of the second series. For each pair we compute the difference between the two values of the measure. The line corresponding to a zero difference separates the plane between pairs for which the first matrix is preferred by the measure, and pairs for which the second one is. We call this line the *discrimination line*.

Our approach is related to the isometrics concept described in [13]. The main difference is our discrimination lines are function of the error distribution, while ROC curves uses TPR and FPR. Moreover, we focus on a single isometrics: the one associated with a zero difference. This allows us to represent several discrimination lines on the same plots. By repeating the same process for different class proportions, or number of classes, we can therefore study if and how the discrimination lines are affected by these parameters.

We now focus on the modeling of classification error distribution. Let us consider a confusion matrix corresponding to a perfect classification, as presented in Table 4. Applying a classifier with lower performance on the same dataset will lead to a matrix diverging only in the distribution of instances in columns taken independently. Indeed, since the dataset is fixed, the class proportions, and hence the distributions inside rows, cannot change (i.e. the $\pi_i$ are constant).

**Table 9:** confusion matrix with controlled errors for a classifier imperfect in all classes.

|  | $C_1$ | ... | $C_j$ | ... | $C_k$ |
|---|---|---|---|---|---|
| $\hat{C}_1$ | $c_1\pi_1$ | ... | $\frac{1-c_j}{k-1}\pi_j$ | ... | $\frac{1-c_k}{k-1}\pi_k$ |
| $\vdots$ | $\vdots$ | $\ddots$ | $\vdots$ | $\vdots$ | $\vdots$ |
| $\hat{C}_i$ | $\frac{1-c_1}{k-1}\pi_1$ | ... | $c_j\pi_j$ | ... | $\frac{1-c_k}{k-1}\pi_k$ |
| $\vdots$ | $\vdots$ | $\vdots$ | $\vdots$ | $\ddots$ | $\vdots$ |
| $\hat{C}_k$ | $\frac{1-c_1}{k-1}\pi_1$ | ... | $\frac{1-c_j}{k-1}\pi_j$ | ... | $c_k\pi_k$ |

For simplicity purposes, we suppose the misclassified instances for some class $C_i$ are uniformly distributed by the classifier on the other estimated classes $\hat{C}_{j\neq i}$. In other words, the perfect classifier correctly puts a proportion $\pi_i$ of the dataset instances in $\hat{C}_i$ and none in $\hat{C}_{j\neq i}$, whereas our imperfect classifier correctly process only a proportion $c_i\pi_i$ ($0 \leq c_i \leq 1$) and incorrectly puts a proportion $\frac{1-c_i}{k-1}\pi_{j\neq i}$ in each other class $\hat{C}_{j\neq i}$, where $1-c_i$ is the accuracy drop for this class. This allows us to control the error level in the confusion matrix, a perfect classification corresponding to $c_i = 1$ for all classes. Table 9 represents the confusion matrix obtained for a $k$-class problem in the case of a classifier undergoing an accuracy drop in all classes.

By using a range of values in $]0;1]$ for $c$, we can generate a series of matrices with decreasing error level. However, comparing pairs of matrices from the same series is fruitless, since it will lead by construction to the same discrimination lines for all measures, when we want to study their differences. We therefore considered two different series: in the first (represented on the $x$ axis), the same accuracy drop $c$ is applied to all classes, whereas in the second ($y$ axis), it is applied only to the first class. In Table 9, the first series corresponds to $c_i = c$ ($\forall i$), and the second to $c_1 = c$ and $c_{i \geq 2} = 1$. We thus expect the accuracy measures to favor the second series, since only its first class is subject to classification errors..

To investigate the sensitivity of the measures to different class proportions values (i.e. $\pi_i$), we generated several pairs of series with controlled class imbalance. In the balanced case, each class represents a proportion $\pi_i = 1/k$ of the instances. We define the completely imbalanced case by defining the 1st class as having twice the number of instances in the 2nd one, which has itself twice the size of the 3rd one, and so on. In other words, $\pi_i = 2^{k-i}/(2^k - 1)$, where the denominator corresponds to the quantity $\sum_{m=0}^{k-1} 2^m$ and allows the $\pi_i$ summing to unity. To control the amount of variation in the class proportion between the balanced and imbalanced cases, we use a multiplicative coefficient $p$ ($0 < p < 1$). Finally, the class proportions are defined as:

$$\pi_i(p) = (1-p)/k + p\, 2^{k-i}/(2^k - 1) \qquad (10)$$

The classes are perfectly balanced for $p=0$ and they become more and more imbalanced as $p$ increases. For instance, a fully imbalanced 5-class datasets will have the following proportions: 0.52, 0.26, 0.13, 0.06 and 0.03, from the 1st to 5th classes, respectively. For each measure, we are now able to plot a discrimination line for each considered value of $p$. This allows us not only to compare several measures for a given $p$ value but also the different discrimination lines of a single measure as a function of $p$.

## 6.2 Error Distribution Sensitivity

We generated matrices for 3 balanced classes ($p=0$) using the methodology described above. Fig. 1 and 2 show the discrimination lines for class-specific and multiclass measures, respectively. Except for TPR, all discrimination lines are located under the $y=x$ line. So, as expected, the measures favor the case where the errors are uniformly distributed in one class ($y$ series) against the case where the errors affect all the classes ($x$ series).

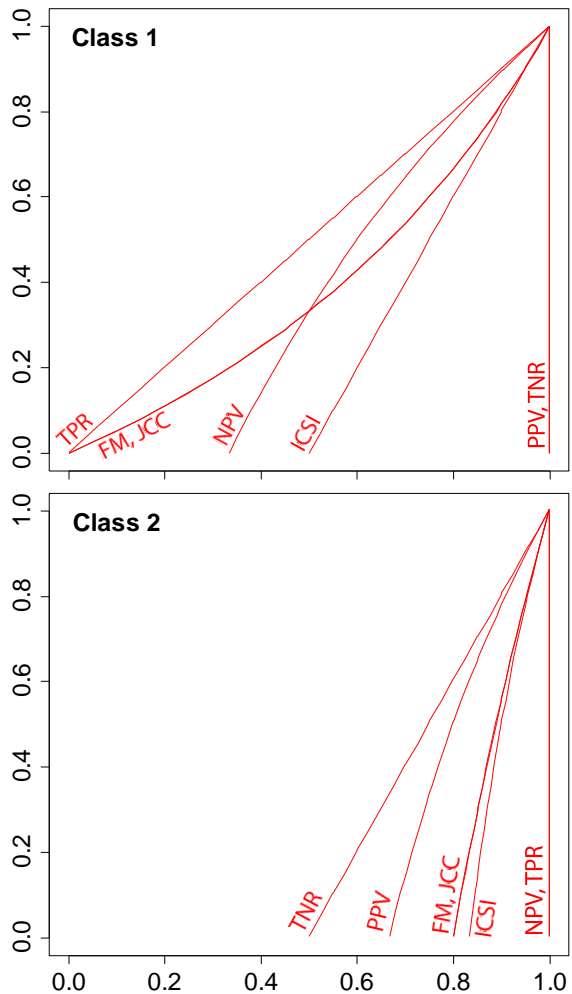

**Figure 1:** Discrimination lines of all class-specific measures for classes 1 (top) and 2 (bottom), for 3 balanced class ($p=0$, $k=3$).

For class-specific measures, we considered first class 1, which is affected by error distribution changes in both series. The discrimination lines are clearly different for all measures. TPR is affected by changes in the distribution of instances only inside the column associated to the considered class. In the case of the first class, these columns are similar on both axes: this explains the $y=x$ discrimination line. The F-measure additionally integrates the PPV value. This explains why it favors the $y$ series matrices. Indeed, the PPV is always greater (or equal) for this series due to the fact errors are present in the first class only. The discrimination line of JCC is exactly similar. NPV does not consider TP, so it is constant for the $y$ series, whereas it decreases for the $x$ series. This is due to the fact more and more errors are added to classes 2 and 3 in these matrices when $p$ increases. This explains why matrices of the $y$ series are largely favored by this measure. PPV and TNR are represented as a vertical line on the extreme right of the plot. According to these measures, the $y$ series matrices are always more accurate. This is due to the fact both measures decrease when the error level increases for the $x$ series ($p_{TN}$ decreases, $p_{FP}$ increases) whereas TNR is constant and PPV decreases less for the $y$ series. Finally, ICSI, which is a linear combination of PPV and TPR, lies in between those measures.

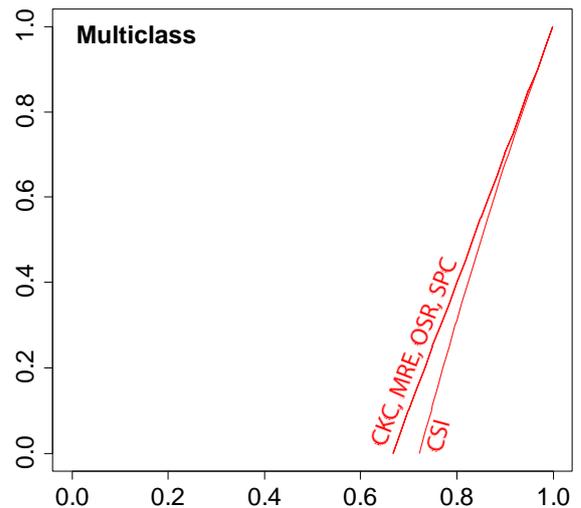

**Figure 2:** Discrimination lines of all multiclass measures, with $p=0$ and $k=3$.

The two other classes undergo similar changes,

so we only report the results for class 2. Unlike class 1, both classes 2 and 3 are affected by errors only in the $x$ series matrices. Consequently, all measures clearly favor the $y$ series matrices, even more than for class 1. The discrimination lines for NPV and TPR take the form of a vertical line on the right of the plot. This is due to the fact both measures decrease only for $x$ series matrices (because of the increase in $p_{FN}$ and $p_{TP}$). The F-measure and JCC are still not discernable. TNR and PPV favor the $y$ series less than the other measures. This is due to the fact that on the one hand $p_{TP}$ decreases only for the $x$ series matrices, but on the other hand $p_{FP}$ and $p_{TN}$ decrease for both series. Finally, ICSI still lies in between PPV and TPR.

Except for CSI the discrimination lines of multiclass measures are identical. We can conclude that for balanced classes ($p=0$) these measures are equivalent. CSI is more sensitive to the type of error we introduced. Indeed it clearly favors the $y$ series more than the other measures.

### 6.3 Class Proportions Distribution Sensitivity

We now study the sensitivity of the on measures on variation in the class proportions. Roughly speaking we observe two different type of behaviors: measures are either sensitive or insensitive to variations in the class proportions distribution. In the first case, the discrimination lines for the different values of $p$ are identical. In the second case, increasing the imbalance leads to lines located on the left of the $p=0$ line. The stronger the imbalance and the more the line is located on the left. This can be explained by the fact the more imbalanced the classes and the more similar the two series of matrices become, dragging the discrimination line closer to the $y=x$ line. Fig. 3 is a typical example of this behavior. It represents the results obtained for the F-measure applied to classes 1 and 3. Note that, like before, JCC and the F-measure have similar discrimination lines.

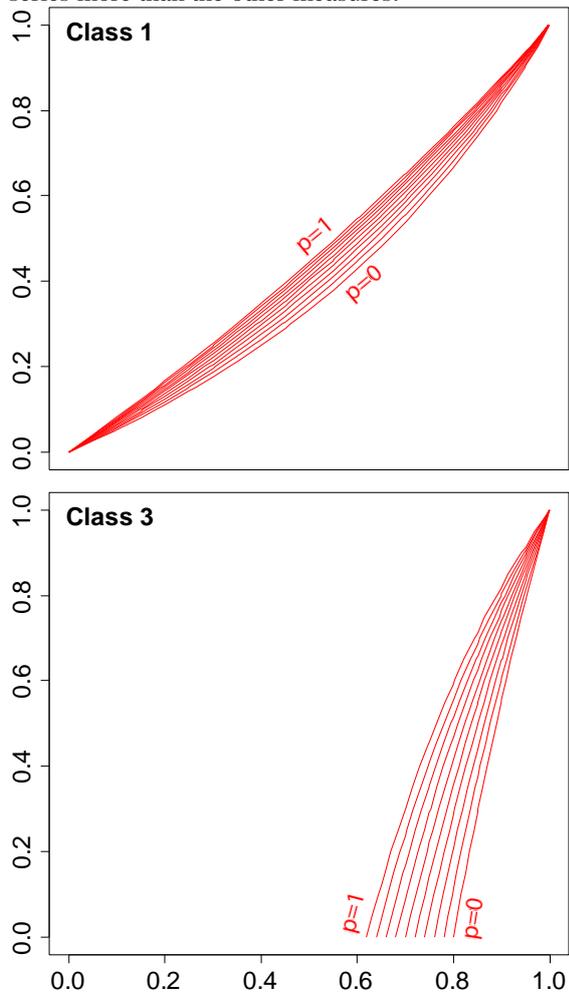

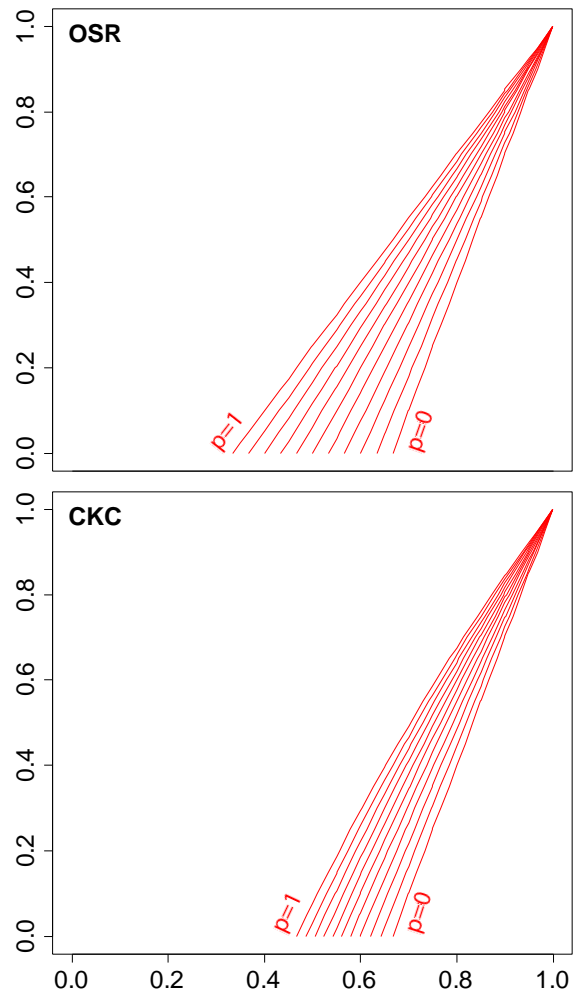

**Figure 3:** Discrimination lines of the F-measure for classes 1 (top) and 3 (bottom), with $k=3$.

**Figure 4:** Discrimination lines of OSR (top) and CKC (bottom), with $k=3$.

Other than the F-measure, measures combining two marginal rates (ICSI, JCC) are sensitive to class proportions changes for all classes. This is not the case for simple marginal rates. TPR and NPV are not sensitive at all for any classes. TNR and PPV present the behavior of measures sensitive to this parameter, but only for classes 2 and 3. As mentioned before, by construction of the considered matrices (the $y$ series has errors only in class 1) they are always higher for the $y$ than the $x$ series, independently of the class proportions. Fig. 4 represents results obtained for the multiclass measures. As previously observed in the balanced case ($p=0$), OSR, SPC and MRE share the same discrimination lines, and this independently of $p$. CKC was matching them for $p=0$, but this is no more the case for imbalanced classes. The plot for CSI (not represented here) is similar but with tighter discrimination lines, indicating it is less sensitive to proportion changes.

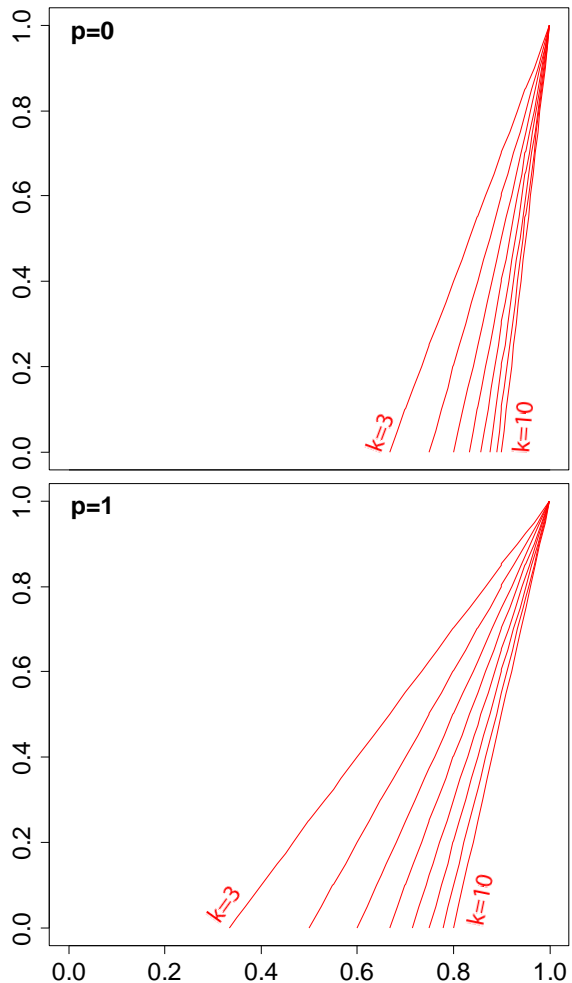

**Figure 5:** Discrimination lines of OSR for $p=0$ (top) and $p=1$ (bottom), with $k \in [3;10]$.

### 6.4 Class Number Sensitivity

We finally focus on the effect of the number of classes on the multiclass measures. Fig. 5 shows the results for OSR applied on matrices with size ranging from 3 to 10, for balanced ($p=0$) and imbalanced ($p=1$) cases. All the measures follow the same behavior. Increasing the number of classes strengthens the preference towards the $y$ series matrices. In other words, having more classes gives more importance to the additional errors contained in the $x$ series matrices. The effect is stronger on the imbalanced matrices. In this case, most of the instances are in the first class, which is the only one similar between the two models, so its dilution has a stronger impact on the measured accuracy.

## 7 DISCUSSION

As shown in the previous sections, measures differ in the way they discriminate different classifiers. However, besides this important aspect, they must also be compared according to several more theoretical traits.

### 7.1 Class Focus

As illustrated in the previous sections, a measure can assess the accuracy for a specific class or over all classes. The former is adapted to situations where one is interested in a given class, or wants to conduct a class-by-class analysis of the classification results.

It is possible to define an overall measure by combining class-specific values measured for all classes, for example by averaging them, like in CSI. However, even if the considered class-specific measure has a clear meaning, it is difficult to give a straightforward interpretation to the resulting overall measure, other than in terms of combination of the class-specific values. Inversely, it is possible to use an overall measure to assess a given class accuracy, by merging all classes except the considered one [2]. In this case, the interpretation is straightforward though, and depends directly on the overall measure.

One generally uses a class-specific measure in order to distinguish classes in terms of importance. This is not possible with most basic overall measures, because they consider all classes to be equally important. Certain more sophisticated measures allow associating a weight to each class, though [7]. However, a more flexible method makes this built-in feature redundant. It consists in associating a weight to each cell in the confusion matrix, and then using a regular (unweighted) overall measure [27]. This method allows distinguishing, in terms of importance, not only classes, but also any possible case of classification error.

### 7.2 Functional Relationships

It is interesting to notice that various combinations of two quantities can be sorted by increasing order, independently from the considered quantities: minimum, harmonic mean, geometric mean, arithmetic mean, quadratic mean, maximum [32]. If the quantities belong to $[0;1]$, we can even put their product at the beginning of the previous list, as the smallest combination. If we consider the presented measures, this means combinations of the same marginal rates have a predefined order for a given classifier. For instance, the sensitivity-precision product will always be smaller than the F-measure (harmonic mean), which in turn will always be smaller than Kulczynski's measure (arithmetic mean). Besides these combinations of TPR and PPV, this also holds for various measures corresponding to combinations of TPR and TNR, not presented here because they are not very popular [20, 33].

More importantly, some of the measures we presented are monotonically related, and this property takes a particular importance in our situation. Indeed, our goal is to sort classifiers depending on their performance on a given data set. If two measures are monotonically related, then the order will be the same for both measures. This makes the F-measure and Jaccard's coefficient similar for classifier comparison, and so are the ICSI and Kulczynski's measure, and of course all measures defined as complements of other measures, such as the FNR. This confirms some of our observations from the previous section: it explains the systematic matching between JCC and the F-measure discrimination lines.

### 7.3 Range

In the classification context, one can consider two extreme situations: perfect classification (i.e. diagonal confusion matrix) and perfect misclassification (i.e. all diagonal elements are zeros). The former should be associated to the upper bound of the accuracy measure, and the latter to its lower bound.

Measure bounds can either be fixed or depend on the processed data. The former is generally considered as a favorable trait, because it allows comparing values measured on different data sets without having to normalize them for scale matters. Moreover, having fixed bounds makes it easier to give an absolute interpretation of the measured features.

In our case, we want to compare classifiers evaluated on the same data. Furthermore, we are interested in their relative accuracies, i.e. we focus only on their relative differences. Consequently, this trait is not necessary. But it turns out most authors normalized their measures in order to give them fixed bounds (usually $[-1;1]$ or $[0;1]$). Note their exact values are of little importance, since any measure defined on a given interval can easily be rescaled to fit another one. Thus, several supposedly different measures are actually the same, but transposed to different scales [34].

### 7.4 Interpretation

Our goal is to compare classifiers on a given dataset, for which all we need is the measured accuracies. In other words, numerical values are enough to assess which classifier is the best on the considered data. But identifying the best classifier is useless if we do not know the criteria underlying this discrimination, i.e. if we are not able to interpret the measure. For instance, being the best in terms of PPV or TPR has a totally different meaning, since these measures focus on type I and II errors, respectively.

Among the measures used in the literature to assess classifiers accuracy, some have been designed analytically, in order to have a clear interpretation (e.g. Jaccard's coefficient [4]). Sometimes, this interpretation is questioned, or different alternatives exist, leading to several related measures (e.g. agreement coefficients). In some other cases, the measure is an *ad hoc* construct, which can be justified by practical constraints or observation, but may lack an actual interpretation (e.g. CSI). Finally, some measures are heterogeneous mixes of other measures, and have no direct meaning (e.g. the combination of OSR and marginal rates described in [35]). They can only be interpreted in terms of the measures forming them, and this is generally considered to be a difficult task.

### 7.5 Correction for Chance

Correcting measures for chance is still an open debate. First, authors disagree on the necessity of this correction, depending on the application context [7, 27]. In our case, we want to generalize the accuracy measured on a sample to the whole population. In other terms, we want to distinguish the proportion of success the algorithm will be able to reproduce on different data from the lucky guesses made on the testing sample, so this correction seems necessary.

Second, authors disagree on the nature of the correction term, as illustrated in our description of agreement coefficients. We can distinguish two kinds of corrections: those depending only on the true class distribution (e.g. Scott's and Maxwell's) and those depending also on the estimated class distribution (e.g. Cohen's and Türk's). The former is of little practical interest for us, because such a measure is linearly related to the OSR (the correction value being the same for every tested algorithm), and would therefore lead to the same ordering of algorithms. This explains the systematic matching observed between the discrimination lines of these measures in the previous section. The latter correction is more relevant, but there is still concern regarding how chance should be modeled. Indeed, lucky guesses depend completely on the algorithm behind the considered classifier. In other words, a very specific model would have to be designed for each algorithm in order to efficiently account for chance, which seems difficult or even impossible.

## 8 CONCLUSION

In this work, we reviewed the main measures used for accuracy assessment, from a specific classification perspective. We consider the case where one wants to compare different classification algorithms by testing them on a given data sample, in order to determine which one will be the best on the sampled population.

We first reviewed and described the most widespread measures, and introduced the notion of discrimination plot to compare their behavior in the context of our specific situation. We considered three factors: changes in the error level, in the class proportions, and in the number of classes. As expected, most measures have a proper way to handle the error factor, although some similarities exist between some of them. The effect of the other factors is more homogeneous: decreasing the number of classes and/or increasing their imbalance tend to lower the importance of the error level for all measures.

We then compared the measure from a more theoretical point of view. In the situation studied here, it turns out several traits of the measures are not relevant to discriminate them. First, all monotonically related measures are similar to us, because they all lead to the same ordering of algorithms. This notably discards a type of chance correction. Second, their range is of little importance, because we are considering relative values. Moreover, a whole subset of measures associating weights to classes can be discarded, because a simpler method allows distinguishing classes in terms of importance while using an unweighted multiclass measure. Concerning chance-correction, it appears it is needed for our purpose; however no existing estimation for chance seems relevant. Finally, complex measures based on the combination of other measures are difficult or impossible to interpret correctly.

Under these conditions, we advise the user to choose the simplest measures, whose interpretation is straightforward. For overall accuracy assessment, the OSR seems to be the most adapted. If the focus has to be made on a specific class, we recommend using both the TPR and PPV, or a meaningful combination such as the F-measure. A weight matrix can be used to specify differences between classes or errors.

We plan to complete this work by focusing on the slightly different case of classifiers with real-valued output. This property allows using additional measures such as the area under the ROC curve and various error measures [20].